\documentclass[conference]{IEEEtran}

\usepackage[utf8]{inputenc}
\usepackage{graphicx}
\usepackage{amsmath, amssymb}
\usepackage{multirow}
\usepackage{float}
\usepackage{subcaption}
\usepackage{url}
\usepackage{booktabs}
\usepackage{lipsum}
\usepackage{array}
\usepackage{hyperref}

\usepackage[backend=biber, style=ieee, sorting=none]{biblatex}
\title{Learning-Based Vision Systems for Semi-Autonomous Forklift Operation in Industrial Warehouse Environments}

\author{
\IEEEauthorblockN{Vamshika Sutar\IEEEauthorrefmark{1}, 
                  Mahek Maheshwari\IEEEauthorrefmark{1}, 
                  Archak Mittal\IEEEauthorrefmark{1}\IEEEauthorrefmark{2}}
                  
\IEEEauthorblockA{\IEEEauthorrefmark{2}Corresponding Author and Assistant Professor }

\IEEEauthorblockA{\IEEEauthorrefmark{1}Department of Civil Engineering, Indian Institute of Technology Bombay, India \\
archak@iitb.ac.in}
}

\date{}

\begin{document}

\maketitle

\begin{abstract}
The automation of material handling in warehouses increasingly relies on robust, low-cost perception systems for forklifts and Automated Guided Vehicles (AGVs). This work presents a vision-based framework for pallet and pallet-hole detection and mapping using a single standard camera. We utilized YOLOv8 and YOLOv11 architectures, enhanced through Optuna-driven hyperparameter optimization and spatial post-processing. An innovative pallet-hole mapping module converts the detections into actionable spatial representations, enabling accurate pallet–hole association for semi-autonomous forklift operation. Experiments on a custom dataset augmented with real warehouse imagery show that YOLOv8 achieves up to 95\% pallet detection  accuracy and 72\% pallet-hole accuracy, while YOLOv11, particularly under optimized configurations, offers superior precision and stable convergence. The results demonstrate the feasibility of a cost-effective, retrofittable visual perception module for forklifts. This study proposes a scalable approach to advancing warehouse automation, promoting safer, economical, and intelligent logistics operations.
\end{abstract}


\section{Introduction}

The warehousing and logistics sector is undergoing a rapid transformation driven by e-commerce growth and persistent labor shortages, with companies citing labor as a top challenge. This has accelerated investment in automation, with companies planning to evaluate Automated Guided Vehicles (AGVs) and Autonomous Mobile Robots (AMRs) in the next few years. As a cornerstone of this shift, AGVs and robotic forklifts are pivotal for automating material handling tasks like pallet transportation, stacking, and retrieval \cite{zhang2023application}. The linchpin of these operations is robust perception,  the ability to accurately detect, localize, and precisely align with pallets and their insertion points (fork holes). This capability is fundamental not only for navigation but also for ensuring the safe and efficient physical interaction required for material handling.

State-of-the-art solutions for this perceptual challenge have traditionally relied on sophisticated sensor suites. Methods using QR codes or high-resolution 3D LiDARs are common but introduce significant drawbacks, including high cost, sensitivity to viewing angles, and limitations in dynamic environments. While advanced vision-based methods like lightweight segmentation networks (LCS-Net) show promise for precise pallet localization using visual data \cite{wu2024pallet}, achieving a balance between high accuracy, real-time inference speed, and generalization across diverse pallet types remains a significant challenge. Furthermore, the prospect of replacing entire fleets of manual forklifts with new, sensor-integrated autonomous models is often financially untenable, highlighting a critical gap in the availability of low-cost, retrofittable perception technologies \cite{faccio2019collaborative}.

To bridge this gap, we propose a vision-based framework for pallet and pallet-hole detection and mapping that relies exclusively on a single standard camera. Our approach leverages the latest advancements in efficient deep learning, specifically YOLOv8 and a next-generation YOLOv11 architecture, to achieve high-precision detection at real-time inference speeds. This design prioritizes cost-effectiveness and ease of integration, allowing for seamless retrofitting onto existing forklift fleets.

Moving beyond mere detection, our second key contribution is a novel pallet-hole mapping module. This component transforms raw, 2D bounding box detections into actionable spatial representations within the vehicle's operational framework. By accurately estimating the insertion points of pallets, this module provides the essential intelligence for AGVs to execute precise navigation, alignment, and successful fork insertion. This end-to-end framework delivers a robust, scalable, and practical perception system designed to accelerate the adoption of automation in real-world warehouse environments.

\section{Related Work}

Research in warehouse automation has advanced through developments in computer vision and deep learning, focusing on enabling precise pallet handling for autonomous systems. Key areas include pallet detection and pallet hole detection, essential for navigation, stacking, and fork alignment. Early methods relied on handcrafted features and geometric models, which struggled in cluttered or dynamic environments. Recent approaches leverage CNN-based detectors, RGB-D sensing, and attention mechanisms to improve accuracy and speed. However, challenges remain in robust hole detection and mapping. The following sections review prior work in these areas, identifying gaps that motivate our proposed YOLO-based solution for integrated pallet and hole detection.

\subsection{Pallet Detection}

Pallet detection is a fundamental capability in warehouse automation, as it enables forklifts and automated guided vehicles (AGVs) to localize load carriers for navigation, stacking, and retrieval. Early research was mainly focused on detecting the overall pallet structure from monocular RGB images using deep learning-based detectors. 

Xiao et al. \autocite{xiao2017pallet} demonstrated pallet recognition and localization from RGB-D contours, but their system was limited to pallet body detection.
Mohamed et al. \autocite{mohamed2020detection} proposed a hybrid pallet detection and localization framework combining Faster R-CNN on RGB images with CNN-based classification of 2D rangefinder data, followed by Kalman filtering. Their system achieved 99.58\% detection accuracy and $\sim$2\,cm localization error, though it did not address pallet hole detection, limiting fork insertion tasks. Kai et al. \autocite{kai2025pallet} proposed a CNN-based front-face shot framework to estimate pallet pose for forklift alignment. Hussain et al. \autocite{hussain2023custom} designed a lightweight CNN for detecting damaged pallet racks, optimized for real-time embedded deployment. While focused on rack integrity, their results highlight the potential of efficient CNNs for pallet-related perception. Keshri et al. \autocite{keshri2023yolox} developed a YOLOX-based approach with a monocular RGB camera, using focal length calibration to estimate pallet coordinates and distances. The system achieved $\sim$70\% accuracy, but again lacked pallet hole detection, constraining automation precision. More recently, Kita et al. \autocite{kita2024pallet360} leveraged $360^\circ$ image projections for pallet detection and 3D pose estimation, attaining $\pm 1^\circ$ yaw error and $\sim$2\,cm position error, making it practical for forklift alignment with pallets. Gann et al. \autocite{gann2024synthetic} investigated pallet detection using synthetic and domain-randomized datasets with YOLOv8, achieving significant improvements in detecting stacked and racked pallets under varied lighting. 

Overall, these studies demonstrate significant progress in pallet detection but remain limited to CNN-based or 2D/3D detection frameworks without explicit pallet hole recognition. In contrast, our work employs advanced YOLO models (YOLOv8, YOLOv11) to achieve both pallet and pallet hole detection, along with pallet-hole mapping for AGV operations, enabling a more complete solution for warehouse automation.

\subsection {Pallet hole detection}

Pallet detection, pallet hole detection, and the reliable mapping of pallet holes to their corresponding pallets are essential for enabling automated forklifts to accurately position forks for load handling. Early research by Cucchiara et al. \autocite{cucchiara2000pallet} introduced a model-based method that combined Hough transforms with virtual corner features to identify pallets and approximate pallet holes, providing a foundation for fork guidance. However, such handcrafted approaches lacked robustness in cluttered and dynamic warehouse environments. Building on advances in sensing, Xiao et al. \autocite{xiao2017pallet} proposed an RGB-D based system that detected pallet structures and holes using geometric constraints, enabling fork insertion but with degraded performance under poor lighting and occlusion. More recently, Wang and Bo \autocite{wang2025pallet} enhanced YOLOv5 with Efficient Channel Attention (ECA) and Ghost-Shuffle Convolution (GSC) to improve pallet hole recognition in RGB-D images, achieving higher accuracy and real-time performance, yet their method did not establish explicit correspondence between detected holes and pallets. Addressing this gap, our approach introduces a centroid- and IoU-based mapping strategy that explicitly links pallet holes to their respective pallets, ensuring robust and precise fork alignment in real-world warehouse environments.

\section{Dataset}

The data pipeline includes a custom dataset from an industrial warehouse focused on pallet and pallet hole detection, and a publicly available multi-class warehouse object detection dataset from Roboflow. Together, these datasets provide a comprehensive basis for training models capable of handling complex real-world scenarios.

\subsection{Pallet and Pallet Hole Detection Dataset}

To enable accurate detection of pallets and their corresponding holes, we assembled a dataset comprising 1,694 images, including 1,523 images sourced from the Roboflow platform and an additional 171 images captured directly within a warehouse environment.

\begin{figure*} 
    \centering
    \begin{subfigure}[b]{0.45\textwidth}
        \centering
        \includegraphics[width=\textwidth]{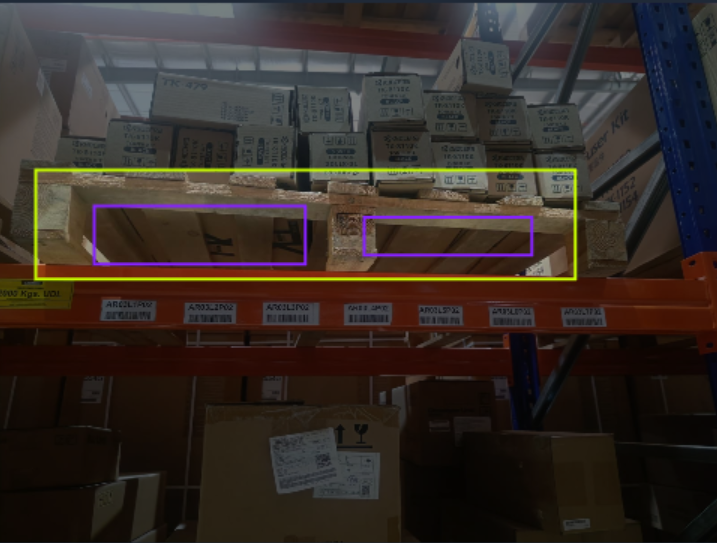}
    \end{subfigure}
    \hfill
    \begin{subfigure}[b]{0.45\textwidth}
        \centering
        \includegraphics[width=\textwidth]{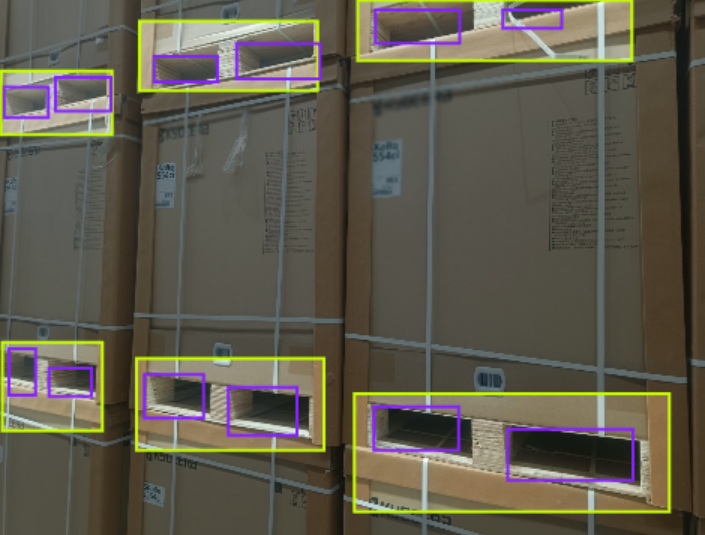}
    \end{subfigure}

    \caption{Annotated images from the dataset.}
    \label{fig:Annotated_images}
\end{figure*}

As shown in Figure~\ref{fig:Annotated_images}, the images were annotated and the dataset was partitioned into training (75\%), validation (10\%), and test (15\%) subsets. To improve the generalization capability of the detection models and to better replicate diverse warehouse conditions, extensive data augmentation techniques were applied. These included horizontal and vertical flipping, random cropping and zooming (up to 19\% zoom), Gaussian blurring (up to 1.1 pixels), and additive noise injection (affecting up to 0.49\% of image pixels).
These data augmentations were essential in simulating real-world operational challenges, reducing model overfitting and increasing robustness in detection performance.

\section{Methodology}
To enable accurate detection of pallets and their corresponding holes in dynamic warehouse environments, we adopted a structured computer vision pipeline leveraging advanced object detection models. 
We employed different versions of the YOLO (You Only Look Once) models due to their balance of speed and accuracy. We explored more earlier and more recent versions of YOLO models and compared the results across different configurations.We tuned key hyperparameters such as batch size, learning rate, and number of epochs to optimize detection accuracy.
Post-training, we used spatial post-processing techniques, including centroid and IoU-based mapping, to align detected pallet holes with the correct pallets.
We automated hyperparameter optimization using the Optuna framework to systematically explore the parameter space and identify optimal configurations. This methodology ensured a robust, high-precision detection system that can be integrated into automated forklifts. 
\subsection{YOLOv8}
Initially, we employed YOLOv8, a state-of-the-art one-stage object detection model developed by Ultralytics \cite{solawetz2024what_is_yolov8}. YOLOv8 directly predicts bounding boxes and class probabilities in a single forward pass, enabling real-time performance. Unlike earlier versions that relied on anchor boxes, YOLOv8 uses an anchor-free detection head, which assigns object detection responsibility based on the center point of the object, simplifying training and reducing hyperparameter sensitivity.

\begin{figure*} 
    \centering
    \includegraphics[width=\textwidth]{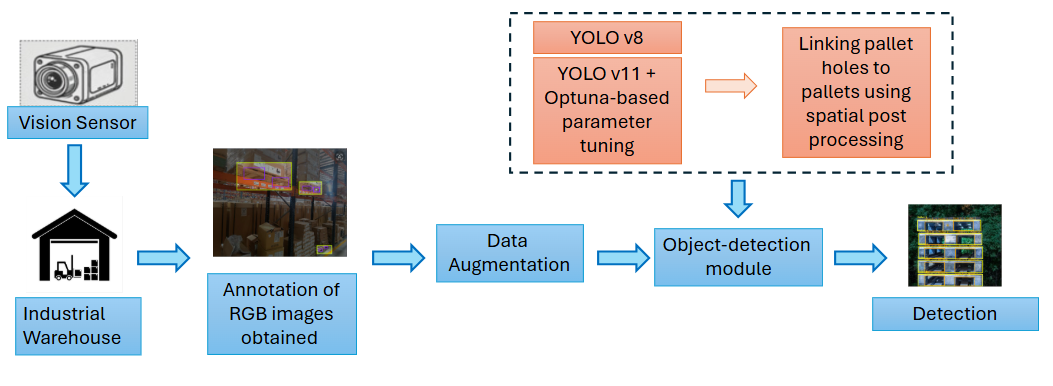} 
    \caption{Pipeline for pallet and pallet-hole detection and association.}
    \label{fig:framework}
\end{figure*}

\subsubsection{YOLOv8 Architecture Overview}
The architecture of YOLOv8 comprises three primary components:
Backbone: The backbone is based on CSP-Darknet (Cross Stage Partial Network), enhanced with C2f modules, which are lightweight convolutional blocks that introduce cross-stage connectivity for efficient feature reuse and gradient flow. These modules contribute to better feature representation while maintaining computational efficiency.
Given an input image $I \in \mathbb{R}^{H \times W \times 3}$, the backbone processes it through successive convolutional layers, generating multi-scale feature maps that encode rich spatial and semantic information.
Neck: YOLOv8 uses a Bidirectional Feature Pyramid Network (BiFPN) structure in the neck. This component fuses features across multiple scales using top-down and bottom-up pathways, allowing better handling of objects at different sizes. The fusion process uses learned attention weights to prioritize semantically stronger features while preserving localization accuracy.
Head: The detection head in YOLOv8 is decoupled, consisting of separate branches for Bounding Box Regression , Objectness Prediction and Class Probability Estimation.
This decoupling improves task-specific learning, reducing interference between regression and classification tasks. Unlike anchor-based models, YOLOv8 uses a center-point-based detection strategy, assigning responsibility to the grid cell containing the object's center.

\subsubsection{Mathematical Formulation}
Given an input image, the object detection model produces an output tensor
$\mathbf{T} \in \mathbb{R}^{S \times S \times (B \cdot (5 + C))}$,  
where $S \times S$ denotes the spatial resolution of the output feature map, determined by the model’s stride (e.g., 8, 16, or 32 pixels). The variable $B$ represents the number of bounding box predictions per grid cell, which is typically set to 1 in anchor-free detection frameworks. 

Each prediction vector consists of five bounding box parameters—center coordinates $(x, y)$, width $(w)$, height $(h)$, and an objectness score—followed by $C$ class probabilities. In this work, $C = 2$, corresponding to the object classes: pallet and pallet-hole.

\subsubsection{Bounding Box Regression Loss}
To optimize the predicted bounding boxes, we employ the Complete Intersection over Union (CIoU) loss, which enhances traditional IoU-based metrics by incorporating spatial and aspect ratio alignment between the predicted and ground-truth boxes. The CIoU loss is defined as:
\[
\mathcal{L}_{\text{CIoU}} = 1 - \text{IoU} + \frac{\rho^2(\mathbf{b}, \mathbf{b}^{gt})}{c^2} + \alpha v
\]
Here, $\text{IoU}$ represents the intersection-over-union between the predicted bounding box $\mathbf{b}$ and the ground-truth box $\mathbf{b}^{gt}$. The term $\rho(\mathbf{b}, \mathbf{b}^{gt})$ denotes the Euclidean distance between the centers of these two boxes, while $c$ is the diagonal length of the smallest enclosing box that encompasses both $\mathbf{b}$ and $\mathbf{b}^{gt}$. The variable $v$ quantifies the similarity of the aspect ratios of the predicted and ground-truth boxes, and $\alpha$ is a dynamic weighting factor that balances this term, defined as:
\[
\alpha = \frac{v}{(1 - \text{IoU}) + v}
\]
This formulation jointly penalizes low overlap, poor center alignment, and dissimilar aspect ratios, thereby leading to more precise and stable bounding box predictions during training. CIoU incorporates spatial overlap (IoU), center point distance, and shape similarity, yielding superior localization accuracy, especially for elongated or irregular objects such as pallet holes.
\subsubsection{ Objectness and Classification Loss}

The objectness score $p_{\text{obj}} \in [0, 1]$ predicts whether an object exists in a given grid cell. Classification and objectness are optimized using Binary Cross Entropy (BCE) loss:
\[
\mathcal{L}_{\text{obj}} = - \left[ y \log(p) + (1 - y) \log(1 - p) \right]
\]
\[
\mathcal{L}_{\text{cls}} = - \sum_{c=1}^{C} y_c \log(p_c)
\]
where $y \in \{0,1\}$ is the ground truth label, and $p$ is the model prediction.
\subsubsection{Total loss}
The overall training objective is a weighted sum of the three components:
\[
\mathcal{L}_{\text{total}} = \lambda_{\text{box}} \cdot \mathcal{L}_{\text{CIoU}} + \lambda_{\text{obj}} \cdot \mathcal{L}_{\text{obj}} + \lambda_{\text{cls}} \cdot \mathcal{L}_{\text{cls}}
\]
where $\lambda_{\text{box}}, \lambda_{\text{obj}}, \lambda_{\text{cls}}$ are tunable hyperparameters controlling the influence of each term.
\subsubsection{Model Training}

To investigate the training dynamics of YOLOv8 for pallet and pallet-hole detection under diverse warehouse conditions, we designed a series of controlled training experiments. These experiments aim to evaluate the impact of data augmentation and batch size on model convergence, generalization, and robustness.
We hypothesize that (i) data augmentation enhances model generalizability across lighting and orientation variations, and (ii) smaller batch sizes, despite slower convergence, may aid generalization by introducing higher gradient noise. Each experiment employs a consistent 75\%–15\%–10\% split of the dataset for training, validation, and testing respectively. The dataset was uploaded on Roboflow platform and accessed via the Roboflow API and was structured in YOLOv8-compatible format.
Model training was conducted using the Ultralytics YOLOv8 framework implemented in Python, with PyTorch as the deep learning backend. We evaluated performance using mean Average Precision at IoU threshold 0.5 (mAP@0.5), Precision, Recall, and loss convergence curves on validation and test sets.
All experiments were conducted using an NVIDIA RTX A6000 GPU, providing substantial computational capacity for high-throughput training and inference. The environment was configured with CUDA version 12.8 and NVIDIA driver version 570.124.06 to ensure optimal compatibility and performance with the Ultralytics YOLOv8 framework.
Each YOLOv8 model was trained for 100 epochs, with a typical training duration of approximately 10–15 minutes per model, owing to efficient hardware utilization and the moderate dataset size. For rigorous evaluation and reproducibility, fixed random seeds were used across all training runs. Performance metrics, including loss curves, mAP scores, precision-recall, and class-wise confusion matrices, were consistently logged via the Ultralytics dashboard, and model checkpoints were systematically saved to preserve experimental states. \\

Three independent YOLOv8 models were trained to isolate the effects of augmentation and batch size. Each model retained identical architecture and hyperparameters unless otherwise stated.Table~\ref{tab:model-configsv8} shows the different model configurations of YOLOv8 trained on the dataset. 


\begin{table*}
\centering
\begin{tabular}{|c|c|c|c|c|c|c|}
\hline
\textbf{Model} & \textbf{Augmentation} & \textbf{Batch Size} & \textbf{Epochs} & \textbf{Initial LR} & \textbf{Scheduler} & \textbf{Optimizer} \\ \hline
Model 1 & Yes & 60 & 100 & 0.005 & Cosine Annealing & SGD with Momentum \\ \hline
Model 2 & No & 60 & 100 & 0.005 & Cosine Annealing & SGD with Momentum \\ \hline
Model 3 & No & 16 & 100 & 0.005 & Cosine Annealing & SGD with Momentum \\ \hline
\end{tabular}
\caption{Baseline Training Configuration for YOLOv8 models}
\label{tab:model-configsv8}
\end{table*}

\begin{table*}
\centering
\begin{tabular}{|l|c|c|c|c|c|l|}
\hline
\textbf{Model} & \textbf{Augmentation} & \textbf{Batch Size} & \textbf{Epochs} & \textbf{Initial LR} & \textbf{Scheduler} & \textbf{Optimizer} \\
\hline
Model 1 & Yes & 60 & 100 & 0.005 & Cosine Annealing & SGD with Momentum \\
Model 2 & Yes & 60 & 150 & 0.005 & Cosine Annealing & SGD with Momentum \\
\hline
\end{tabular}
\caption{Baseline Training Configurations for YOLOv11 Models}
\label{tab:model-configsv11}
\end{table*}

Model~1 serves as the baseline, trained with a comprehensive data augmentation pipeline to improve generalization under varying pallet orientations, lighting, and background noise. 

We applied these augmentations using the built-in Ultralytics augmentation modules. The model was trained for 100 epochs with a batch size of 60, an initial learning rate of 0.005, using Stochastic Gradient Descent (SGD) with momentum. Learning rate scheduling followed a cosine annealing strategy.
To isolate the effect of data augmentation, Model 2 was trained with the same architecture, optimizer, and learning rate schedule as Model 1, but with augmentation disabled. This model uses a large batch size of 60 to match Model 1, ensuring that the only variable being tested is the presence or absence of augmentation.
We designed Model 3 to explore the impact of reduced batch size on training stability and generalization,  in the absence of augmentation. This configuration uses a batch size of 16 while keeping all other hyperparameters identical to Models 1 and 2. Smaller batch sizes can increase gradient variance, potentially acting as an implicit regularizer.

By comparing these configurations, we aim to uncover how YOLOv8’s performance varies under different training regimes, especially in challenging real-world detection scenarios such as cluttered warehouse environments with occlusions and lighting inconsistencies.

\subsection{YOLOv11}
In pursuit of enhanced detection robustnes and spatial precision beyond the capabilities of YOLOv8, We adopted YOLOv11, a next-generation one-stage object detector that incorporates architectural improvements to enhance detection accuracy and bounding box localization. YOLOv11 incorporates a series of architectural innovations and advanced loss function formulations that collectively address several limitations observed in YOLOv8 \cite{ultralytics2025yolov8_vs_yolov11}, particularly in complex detection scenarios such as pallet holes in cluttered warehouse environments. These objects often present challenges due to their small size, frequent occlusions, and visual similarity to shadows or structural elements.

\subsubsection{YOLOv11 Architecture Overview}

\paragraph{Backbone}
YOLOv11 retains the foundational Cross Stage Partial (CSP) backbone architecture from YOLOv8 but introduces substantial enhancements to residual aggregation and the C2f (Cross Stage Partial with Fast convolution) modules. These refinements deepen the network’s capacity to learn richer, more discriminative feature representations while maintaining computational efficiency. Enhanced residual aggregation improves gradient propagation, mitigating vanishing gradient issues in very deep networks.

The backbone processes the input image $\mathbf{I} \in \mathbb{R}^{H \times W \times 3}$ through deeper convolutional hierarchies to produce multi-scale feature maps with improved sensitivity to fine-grained spatial cues. This improved design allows YOLOv11 to better differentiate visually similar classes (e.g., pallet holes versus shadows), surpassing the feature extraction capacity of YOLOv8.

\paragraph{Neck}
The neck architecture builds upon the Bi-directional Feature Pyramid Network (BiFPN) from YOLOv8 but integrates learned attention weights for inter-scale feature fusion. Unlike standard BiFPN, which treats all feature levels uniformly, YOLOv11 adaptively weights feature maps based on semantic strength and localization consistency. This attention-guided fusion improves robustness to scale variance and occlusion, critical for detecting pallets and holes at varying distances and perspectives.

\paragraph{Head}
YOLOv11 employs a fully decoupled detection head, in which classification and bounding box regression are handled by separate parallel branches. This architecture, unlike the partially decoupled head in YOLOv8, reduces task interference and enables more specialized feature learning, thereby improving convergence and detection accuracy. In addition, YOLOv11 maintains an anchor-free, center-based detection framework, allowing the model to directly predict object locations without reliance on predefined anchor boxes. This design choice enhances generalization and has been shown to yield more robust performance in complex detection scenarios \cite{ultralyticsYOLO11Docs}.

\subsubsection{Mathematical Formulation}

The prediction tensor output remains consistent with YOLOv8:
\[
\mathbf{T} \in \mathbb{R}^{S \times S \times \left(B \cdot (5 + C)\right)}
\]
where:
\begin{itemize}
    \item $S \times S$ is the grid size determined by feature stride,
    \item $B = 1$ is the number of bounding boxes per grid cell (anchor-free),
    \item $C = 2$ denotes the classes: \texttt{pallet}, \texttt{pallet-hole}.
\end{itemize}

Each output vector comprises:
\[
(x, y, w, h, \text{objectness}, \text{class probabilities})
\]

\subsubsection{Loss Functions}

\paragraph{Bounding Box Regression.}
The YOLOv11 framework employs a geometry-aware bounding box regression objective based on the Complete Intersection over Union (CIoU) formulation, consistent with previous Ultralytics implementations. The CIoU loss enhances bounding box localization by jointly optimizing the overlap area, center point distance, and aspect ratio consistency between predicted and ground-truth boxes.

\paragraph{Objectness and Classification Loss.}
For object confidence and class probability estimation, YOLOv11 retains the Binary Cross-Entropy (BCE) formulation utilized in prior YOLO versions. The objectness branch employs BCE to maintain a robust separation between foreground and background predictions, while the classification branch can optionally leverage focal or varifocal weighting to address class imbalance and emphasize high-confidence samples. These losses are parameterized within the same loss module, ensuring modularity and configurability across different tasks. The official implementation also incorporates the Distribution Focal Loss (DFL) for coordinate refinement, representing bounding box coordinates as discrete distributions to improve sub-pixel localization precision.

\paragraph{Total Loss.}
The overall training objective in YOLOv11 is formulated as a weighted sum of the above components:
\[
\mathcal{L}_{\text{total}} =
\lambda_{\text{box}}\mathcal{L}_{\text{box}} +
\lambda_{\text{obj}}\mathcal{L}_{\text{obj}} +
\lambda_{\text{cls}}\mathcal{L}_{\text{cls}} +
\lambda_{\text{dfl}}\mathcal{L}_{\text{dfl}},
\]
where $\mathcal{L}_{\text{box}}$ denotes the CIoU-based bounding box regression loss, $\mathcal{L}_{\text{obj}}$ and $\mathcal{L}_{\text{cls}}$ correspond to the BCE-based objectness and classification losses, respectively, and $\mathcal{L}_{\text{dfl}}$ represents the Distribution Focal Loss. The weighting coefficients $\lambda$ are empirically tuned to balance the contribution of each term during training.

\subsubsection{Model Training and Evaluation}

For a fair comparison, both YOLOv8 and YOLOv11 were trained under identical experimental settings. The dataset was divided into 75\% for training, 15\% for validation, and 10\% for testing, with annotations provided in the YOLO format exported via Roboflow. All models were trained on an NVIDIA RTX A6000 GPU using PyTorch with CUDA, employing the official Ultralytics YOLOv11 implementation. The training process was conducted for 100–150 epochs depending on the configuration. Table~\ref{tab:model-configsv11} shows the training configurations of YOLO v11 models we trained.

\subsubsection{Parameter Tuning with Optuna}

\paragraph{Optimization Objective.}
We employed Optuna to maximize mAP@0.5:0.95 by tuning:
\begin{itemize}
    \item $\lambda_{\text{cls}}$: classification loss weight,
    \item $\lambda_{\text{box}}$: box regression loss weight,
    \item $\lambda_{\text{iou}}$: IoU loss weight,
    \item $lr_0$: initial learning rate.
\end{itemize}

\paragraph{Search Strategy.}
We employed Optuna for hyperparameter optimization, which leverages the Tree-structured Parzen Estimator (TPE) sampler for adaptive search and the Median Pruner to early-stop trials exhibiting sub-median performance. A total of 20 trials were executed.

\paragraph{Early Stopping: MedianPruner.}
To reduce computation, Optuna’s \texttt{MedianPruner} was employed:
\begin{itemize}
    \item Trials are pruned if their intermediate mAP@0.5:0.95 after 5 epochs falls below the median of ongoing trials.
\end{itemize}

\paragraph{Constants During Tuning.}
\begin{itemize}
    \item Batch size:60
    \item Image resolution: 640 × 640
    \item Optimizer: SGD with momentum
    \item Learning rate scheduler: Cosine Annealing
\end{itemize}

This efficient tuning process enabled effective exploration of the search space and improved final performance without incurring excessive computational costs.

\subsection{Post-processing: Linking Pallets and Pallet Holes}

A critical step in our pipeline is the association of detected pallets with their respective pallet holes. To achieve this, we employed two complementary post-processing strategies: the centroid-based method and the Intersection over Union (IoU) method. 

\begin{figure}
    \centering
    \begin{subfigure}[b]{0.45\textwidth}
        \centering
        \includegraphics[width=\textwidth]{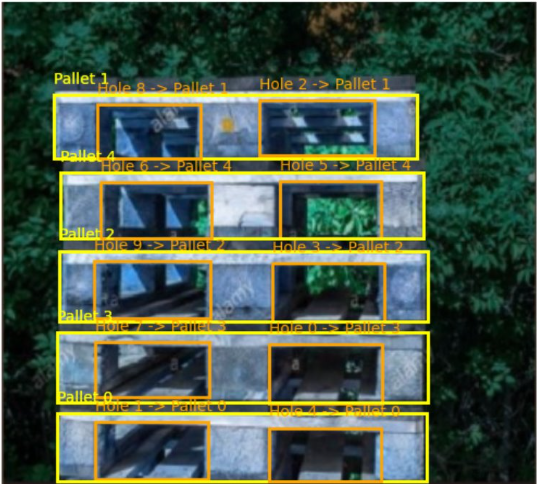}
        \caption{Centroid-based Association}
          \label{fig:centroid}
    \end{subfigure}
    \hfill
    \begin{subfigure}[b]{0.45\textwidth}
        \centering
    \includegraphics[width=\textwidth]{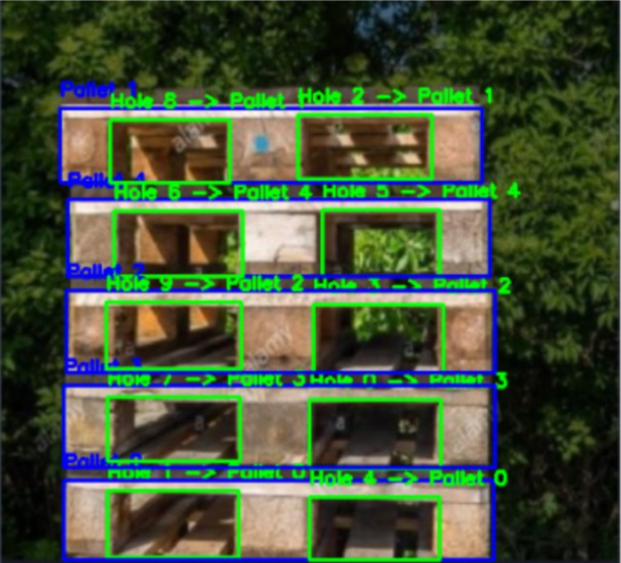}
        \caption{IoU-based Association}
        \label{fig:IoU}
    \end{subfigure}

    \caption{Spatial association techniques to link pallet holes with respective pallets.}
    \label{fig:two_images}
\end{figure}

\paragraph{Centroid-based Association}
For each detected pallet hole $H_i$, we compute its centroid coordinates as
\begin{equation}
C(H_i) = \left( \frac{x_{1}^{h} + x_{2}^{h}}{2}, \, \frac{y_{1}^{h} + y_{2}^{h}}{2} \right),
\end{equation}
where $(x_{1}^{h}, y_{1}^{h})$ and $(x_{2}^{h}, y_{2}^{h})$ denote the top-left and bottom-right corners of the bounding box of $H_i$.  
The centroid $C(H_i)$ is then compared with the bounding boxes of all detected pallets $P_j$. A pallet hole $H_i$ is assigned to pallet $P_j$ if
\begin{equation}
(x_{1}^{p} \leq C_x(H_i) \leq x_{2}^{p}) \quad \land \quad (y_{1}^{p} \leq C_y(H_i) \leq y_{2}^{p}),
\end{equation}
where $(x_{1}^{p}, y_{1}^{p})$ and $(x_{2}^{p}, y_{2}^{p})$ are the bounding box corners of pallet $P_j$.  
As shown in Figure~\ref{fig:centroid} this criterion ensures that each pallet hole is associated with the pallet whose bounding region contains its centroid.

\paragraph{IoU-based Association}
In addition, we utilize the Intersection over Union (IoU) metric to capture the degree of overlap between a pallet hole $H_i$ and a pallet $P_j$. The IoU is defined as
\begin{equation}
\text{IoU}(H_i, P_j) = \frac{|B(H_i) \cap B(P_j)|}{|B(H_i) \cup B(P_j)|},
\end{equation}
where $B(\cdot)$ denotes the bounding box region.  
As shown in Figure~\ref{fig:IoU}, a pallet hole is linked to the pallet $P_j$ that yields the highest IoU value, provided that
\begin{equation}
\text{IoU}(H_i, P_j) \geq \tau,
\end{equation}
with $\tau$ being a predefined threshold (empirically set to balance precision and recall).

\section{Results}

\subsection{YOLOv8 Results}

To assess the detection capabilities of the three proposed YOLOv8-based models, we report a comprehensive set of metrics spanning accuracy, F1 score, and learning performance. The models were evaluated on their ability to detect two key object classes: pallets and pallet holes.

Table~\ref{tab:yolov8-performance} presents class-wise accuracy and F1 scores derived from the normalized confusion matrices and confidence-F1 curves. Notably, Model 3 demonstrates the best overall balance, with a pallet accuracy of 95\% and a pallet hole accuracy of 72\%, alongside a pallet F1 score of 0.93 and pallet hole F1 of 0.62. This indicates its superior generalization across both classes.

Model 1 shows the highest pallet accuracy (97\%) but underperforms in detecting pallet holes. Model 2 records the weakest hole detection (64\%) and lowest hole F1 score (0.55), likely due to lack of augmentation. Model 3 strikes a good trade-off, offering the best hole F1 performance and competitive accuracy, supporting its robustness for real-world deployment.

Model 3 demonstrates improved background separation, reducing misclassification of Class 1 as background. However, some residual confusion remains, largely due to the visual similarity of holes to background structures when captured from distance.

Table~\ref{tab:yolov8-results} captures core performance metrics logged during training and validation. It shows the evolution of bounding box regression loss (\texttt{box\_loss}), classification loss (\texttt{cls\_loss}), and Distribution Focal Loss (\texttt{dfl\_loss}), along with end-task metrics such as precision, recall, and mean Average Precision at IoU thresholds 0.5 and 0.5:0.95. Although Model 1 had the lowest training losses, its validation losses are marginally higher, indicating slight overfitting. In contrast, Model 3 records the best precision, recall, and mAP metrics across the board, despite slightly higher training losses—signaling its superior generalization capabilities. The addition of data augmentation in Model 1 led to slightly improved robustness but did not translate to better performance on pallet-hole detection.

\begin{table*}
\centering
\caption{Performance Comparison and Implementation Results of YOLOv8 Models}
\begin{subtable}{\textwidth}
    \centering
    \begin{tabular}{|l|c|c|c|c|c|}
    \hline
    \textbf{Model} & \textbf{Epochs} & \textbf{Pallet Accuracy} & \textbf{Hole Accuracy} & \textbf{Pallet F1 Score} & \textbf{Hole F1 Score} \\
    \hline
    Model 1 & 100 & 97\% & 71\% & 0.920 & 0.560 \\
    Model 2 & 100 & 95\%          & 64\%          & 0.910 & 0.550 \\
    Model 3 & 100 & 95\%          & 72\%          & 0.930 & 0.620 \\
    \hline
    \end{tabular}
    \vspace{4pt}
    \caption{Performance Comparison of YOLOv8 Models on Pallet and Pallet Hole Detection}
    \label{tab:yolov8-performance}
\end{subtable}

\vspace{1em} 

\begin{subtable}{\textwidth}
    \centering
    \resizebox{\textwidth}{!}{%
    \begin{tabular}{|l|l|c|c|c|c|c|c|c|c|c|c|c|}
    \hline
    \textbf{Model} & \textbf{Params} & \textbf{Epoch} & \textbf{Train Box Loss} & \textbf{Train Cls Loss} & \textbf{Train DFL Loss} & \textbf{Precision} & \textbf{Recall} & \textbf{mAP@0.5} & \textbf{mAP@0.5:0.95} & \textbf{Val Box Loss} & \textbf{Val Cls Loss} & \textbf{Val DFL Loss} \\
    \hline
    Model 1 & v8\_aug\_60   & 100 & 1.038 & 0.543 & 1.046 & 0.774 & 0.797 & 0.743 & 0.490 & 1.427 & 0.760 & 1.254 \\
    Model 2 & v8\_noaug\_60 & 100 & 1.068         & 0.575          & 1.061          & 0.811 & 0.801 & 0.780 & 0.506 & 1.436 & 0.722 & 1.247 \\
    Model 3 & v8\_noaug\_16 & 100 & \textbf{1.138}         & \textbf{0.683}        & \textbf{1.149}        & \textbf{0.814} & \textbf{0.812} & \textbf{0.795} & \textbf{0.516} & \textbf{1.400} & \textbf{0.697} & 1.256 \\
    \hline
    \end{tabular}%
    }
    \vspace{3pt}
 \caption{YOLOv8 Implementation Results}
 \label{tab:yolov8-results}
\end{subtable}
\end{table*}


\subsection{YOLOv11 Results}

This section presents the performance evaluation of YOLOv11 for pallet and pallet-hole detection. The experiments include comparisons across training durations and additional tuning to enhance performance.

\subsubsection{Baseline Comparison: Epochs = 100 vs. 150}





Table~\ref{tab:yolov11_results} captures the performance comparison between YOLOv11 models trained for 100 and 150 epochs. Model 5, trained for 150 epochs, demonstrates marginal improvements in key metrics such as mAP@0.5 and mAP@0.5:0.95, indicating best overall localization performance. Model 4 (100 epochs) achieves higher recall , suggesting that it was more sensitive in identifying true positives, possibly at the expense of slightly higher false positives compared to Model 5.

Training losses for box, classification, and DFL components are consistently lower in Model 5, reflecting more stable convergence with extended training. However, validation losses (particularly box loss and DFL loss) are slightly higher in Model 5, which may indicate mild overfitting or diminishing returns from prolonged training without further regularization.


Despite persistent underperformance in Class 1 (pallet holes) due to visual confusion and dense instance layouts, YOLOv11’s improved architecture, with refined feature fusion and task-specific decoupled heads, enhances spatial awareness and classification focus, helping to mitigate these challenges more effectively than earlier YOLO versions.

While the improvements between 100 and 150 epochs are incremental, the results validate YOLOv11's robustness and responsiveness to extended training. The minor performance gains from longer training suggest that for time-sensitive deployment scenarios, 100-epoch models may offer a practical balance between efficiency and accuracy, while 150-epoch models slightly enhance precision for high-reliability applications.

\begin{table*}
\centering
\caption{Performance Comparison and Implementation Results of YOLOv11 Models}
\begin{subtable}{\textwidth}
    \centering
    \begin{tabular}{|l|c|c|c|c|c|}
    \hline
    \textbf{Model} & \textbf{Epochs} & \textbf{Pallet Accuracy} & \textbf{Hole Accuracy} & \textbf{Pallet F1 Score} & \textbf{Hole F1 Score} \\
    \hline
    Model 1 & 100 & 94\% & 65\% & 0.899 & 0.520 \\
    Model 2 & 150 & 96\% & 68\% & 0.919 & 0.533 \\
    \hline
    \end{tabular}
    \vspace{4pt}
    \caption{Performance Comparison of YOLOv11 Models on Pallet and Pallet Hole Detection}
    \label{tab:yolov11_comparison}
\end{subtable}

\vspace{1em} 

\begin{subtable}{\textwidth}
    \centering
    \resizebox{\textwidth}{!}{%
    \begin{tabular}{|l|l|c|c|c|c|c|c|c|c|c|c|c|}
    \hline
    \textbf{Model} & \textbf{Params} & \textbf{Epoch} & \textbf{Train Box Loss} & \textbf{Train Cls Loss} & \textbf{Train DFL Loss} & \textbf{Precision} & \textbf{Recall} & \textbf{mAP@0.5} & \textbf{mAP@0.5:0.95} & \textbf{Val Box Loss} & \textbf{Val Cls Loss} & \textbf{Val DFL Loss} \\
    \hline
    Model 4 & v11\_aug\_60 & 100 & 1.127 & 0.601 & 1.083 & 0.767 & 0.799 & 0.741 & 0.483 & 1.428 & 0.756 & 1.251 \\
    Model 5 & v11\_aug\_60 & 150 & 1.023 & 0.544 & 1.044 & 0.773 & 0.795 & 0.744 & 0.485 & 1.454 & 0.758 & 1.275 \\
    Trial 14 & v11\_aug\_60 & 150 & \textbf{0.017} & \textbf{0.218} & \textbf{1.057} & \textbf{0.823} & \textbf{0.816} & \textbf{0.788} & \textbf{0.510} & \textbf{0.023} & \textbf{0.262} & \textbf{1.241} \\
    \hline
    \end{tabular}%
    }
    \vspace{3pt}
    \caption{YOLOv11 Implementation Results}
    \label{tab:yolov11_results}
\end{subtable}
\end{table*}

\subsubsection{Parameter Tuning Results}

This section discusses about the results of parameter tuning on the best YOLOv11 model, model, model trained with 150 epochs, and the results for the best trial, Trial 14 are shown in the Table ~\ref{tab:yolov11_results}


Under the 150-epoch setting, the model had more training time to converge, allowing the optimizer to explore a broader range of parameter interactions. The best-performing trial was Trial 14, which significantly improved mAP scores compared to the previous best performing model (Model 5).


\textbf{Best Trial: Trial 14}

\medskip

\noindent Hyperparameters:
\begin{align*}
    \lambda_{\text{cls}} &= 0.1870 \\
    \lambda_{\text{box}} &= 0.1167 \\
    \lambda_{\text{iou}} &= 0.2690 \\
    lr_0 &= 0.0159
\end{align*}

This configuration achieved notably better bounding box alignment and classification consistency. The moderate weights for classification and IoU losses suggest a balanced contribution from spatial localization and class discrimination. The tuned model demonstrated enhanced precision and overall mAP, making it suitable for deployment scenarios requiring high reliability and accuracy.







These results demonstrate that even within constrained training durations, YOLOv11 responds well to careful hyperparameter tuning. Optuna’s pruning strategy and adaptive sampling enabled efficient search through the high-dimensional parameter space, resulting in performance gains with minimal added training cost.

\section{Discussion}

This study presents a comprehensive approach to build an 2D object detection model for automated forklifts, an object detection pipeline for pallet and pallet holes using simple camera captured Front Person View (FPV) images. 

The core technical contribution of this work lies in the implementation and comparative analysis of YOLO-based object detection models, particularly YOLOv8 and YOLOv11, for detecting pallets and pallet holes, which are critical for precise leg insertion in material handling. Across the experiments, YOLOv8 Model 3 emerged with the highest \textit{mAP@50--95}, demonstrating strong overall generalization and localization accuracy. However, YOLOv11 Model 5, despite a marginally lower \textit{mAP@50--95} (0.48455), exhibited superior training dynamics with consistently lower loss values most notably in bounding box regression. While \textit{mAP} remains the primary evaluation metric, YOLOv11's design promises better scalability for more complex detection scenarios, and its improved precision reinforces its readiness for high-confidence applications. With further tuning and larger datasets, YOLOv11 is positioned to surpass earlier variants in both detection fidelity and robustness.

We further augmented detection accuracy through data augmentation, Optuna-based hyperparameter optimization, and spatial post-processing methods like IoU and centroid-based mapping. The best-performing YOLOv11 configuration was obtained via Optuna-based hyperparameter tuning under a 150-epoch regime, with Trial 14 yielding the most promising results among all YOLO v11 models on the test set. 

In contrast to pallets, the detection of pallet holes presents a significantly more difficult problem, even after iterative improvements to the model. One of the primary challenges lies in their visual ambiguity: pallet holes typically appear as dark, rectangular hollow regions, which strongly resemble shadows, gaps, or other background structures when observed from oblique viewpoints or at larger distances. This similarity in low-level visual cues makes it difficult for the detector to reliably distinguish pallet holes from irrelevant background features. Moreover, the structural characteristics of pallet holes introduce additional complexity. Unlike pallets, which are relatively large and well-defined objects, pallet holes are smaller sub-regions embedded within pallets, and their detection often requires fine-grained spatial resolution to avoid misclassification. 

Overall, this study demonstrates a practical and scalable pathway toward intelligent warehouse automation. The combination of centralized perception, teleoperation, and advanced deep learning techniques offers a compelling alternative to high-cost fully autonomous systems, with the potential to significantly improve throughput, reduce labor costs, and increase operational safety in industrial logistics.


\section{Limitations and Future Work}
While our system shows strong results for semi-autonomous material handling, there are some important limitations that affect its performance in real-world warehouse environments.
One key challenge is the misclassification of background objects when images are taken from greater distances. Structures like dark shelves or rectangular shadows can look very similar to pallet holes camera images. Because the resolution and detail of objects reduce at a distance, the model sometimes makes incorrect predictions. 
The accuracy of the model also depends on lighting conditions. Shadows, glares, or poor lighting can make it hard for the model to clearly detect pallet holes, which often rely on visual contrast with the floor. 
Another limitation is that the system struggles with occlusion. In crowded or busy warehouse areas, pallets or pallet holes may be partially hidden from view. This can lead to missed or incorrect detections. The post-processing methods we use, such as centroid and IoU-based matching, work well when bounding boxes are accurate, but can fail when objects overlap or are not fully visible. These methods also do not use temporal information across frames, which could improve detection stability.

Future work can focus on exploring more advanced vision models beyond YOLOv11, particularly transformer-based architectures that capture global context and reduce misclassification at greater distances. Instead of relying solely on bounding box detection, shape-aware segmentation methods could be employed to more accurately identify pallet holes, even under occlusion or low-contrast conditions. Incorporating these techniques would help improve robustness in real-world warehouse environments where lighting, distance, and clutter present significant challenges.

\printbibliography
\end{document}